# Advancing Efficient Brain Tumor Multi-Class Classification -- New Insights from the Vision Mamba Model in Transfer Learning

Yinyi Lai, Anbo Cao, Yuan Gao, Jiaqi Shang, Zongyu Li, Jia Guo

***Abstract*—Early and accurate diagnosis of brain tumors is crucial for improving patient survival rates. However, the detection and classification of brain tumors are challenging due to their diverse types and complex morphological characteristics. This study investigates the application of pre-trained models for brain tumor classification, with a particular focus on deploying the Mamba model. We fine-tuned several mainstream transfer learning models and applied them to the multi-class classification of brain tumors. By comparing these models to those trained from scratch, we demonstrated the significant advantages of transfer learning, especially in the medical imaging field, where annotated data is often limited. Notably, we introduced the Vision Mamba (Vim), a novel network architecture, and applied it for the first time in brain tumor classification, achieving exceptional classification accuracy. Experimental results indicate that the Vim model achieved 100% classification accuracy on an independent test set, emphasizing its potential for tumor classification tasks. These findings underscore the effectiveness of transfer learning in brain tumor classification and reveal that, compared to existing state-of-the-art models, the Vim model is lightweight, efficient, and highly accurate, offering a new perspective for clinical applications. Furthermore, the framework proposed in this study for brain tumor classification, based on transfer learning and the Vision Mamba model, is broadly applicable to other medical imaging classification problems.***

## I. INTRODUCTION

BRAIN tumors present significant challenges in medical diagnosis due to their diverse types and complex morphological characteristics. There are currently over 120 known types of brain and central nervous system (CNS) tumors, ranging from benign to malignant, which makes detection and classification highly challenging. This complexity significantly complicates diagnosis, and the low survival rates associated with brain tumors further underscore their severity [1]. Malignant brain tumors can rapidly compromise a patient's health due to their invasive nature, making early and accurate diagnosis critical for improving survival outcomes [2].

Magnetic Resonance Imaging (MRI) and Computed Tomography (CT) are the primary imaging tools used for diagnosing brain tumors. MRI is preferred for its ability to generate high-resolution images of brain tissue while being non-invasive and avoiding ionizing radiation [3]. MRI can produce detailed images using different sequences, such as T1-weighted, T2-weighted, and FLAIR, which are instrumental in depicting soft tissue structures and detecting and characterizing brain tumors [4][5]. Although CT scans have their uses, particularly in emergency settings, their reliance on ionizing radiation limits their application when frequent imaging is required.

Recent advances in deep learning have revolutionized medical image analysis, significantly enhancing the capacity to tackle complex tasks like brain tumor classification [6]. Convolutional Neural Networks (CNNs), a key part of deep learning, have substantially improved image classification accuracy by automatically learning hierarchical features directly from raw image data [7]. This approach minimizes dependence on manual feature extraction, allowing models to excel in complex tasks. To capture long-range dependencies, Transformer architectures have been introduced, which have demonstrated excellent performance in visual tasks by capturing global relationships within image data [8]. However, traditional Transformers have quadratic attention complexity, which poses substantial computational and memory challenges. This makes them inefficient, particularly for large-scale or 3D medical data, reducing their practicality in clinical settings [9].

Recent research has generated significant interest in State Space Models (SSMs), which are highly efficient for processing long sequences due to their convolutional and near-linear computational characteristics [10]. One notable advancement is the Mamba model [11], which incorporates time-varying parameters into the SSM framework and utilizes a hardware-aware algorithm, achieving efficient training and inference without relying on attention mechanisms. This results in subquadratic computational complexity and linear memory complexity [12]. Mamba's scalability makes it a promising alternative to Transformers, particularly in language modeling. Additionally, Mamba-related models have been successfully introduced in visual processing, demonstrating superior performance in image classification, lesion segmentation, and modality transformation tasks compared to traditional CNNs and some Transformer-based networks.

Despite these advancements, current research reveals certain limitations. Deep learning models generally have a significant demand for data, and in medical imaging, high-resolution annotated datasets are often difficult to obtain. Moreover, existing specialized models exhibit poor generalization across different datasets and may underperform on rare or atypical cases. These models also have high computational costs associated with training [13].

This study aims to address these challenges by exploring the performance of pre-trained models in brain tumor classification tasks, specifically focusing on the Mamba model, which has not yet been applied to this field. The main motivation is to evaluate the effectiveness of the Mamba model in distinguishing between different types of brain tumors and to investigate its potential to improve classification accuracy. By leveraging pre-trained models and fine-tuning, this study aims to enhance diagnostic outcomes and provide new insights into medical image analysis. We provide a comprehensive analysis of the Mamba model's performance and compare it to existing models, demonstrating the efficiency and lightweight characteristics of the Mamba model in processing complex medical image data. This provides a new perspective and potential solutions for the challenging task of brain tumor classification.

The contributions of this paper are as follows:
- Fine-Tuning of Mainstream Models: We fine-tuned several mainstream transfer learning models for brain tumor classification, including Vision Transformer, Swin Transformer, EfficientNet-B0, Inception-V3, and ResNet50.
- Transfer Learning Benefits: Transfer learning, particularly with pre-trained models, is advantageous when working with limited labeled medical data. We compared models trained from scratch with those using transfer learning to demonstrate the clear benefits of pre-trained models.
- Introduction of Vision Mamba: We introduced the Vision Mamba (Vim), a novel network architecture based on Mamba and applied it for the first time to a multi-class brain tumor classification task. Vim achieved outstanding accuracy, further emphasizing the superior capabilities of the Mamba architecture.

These results highlight the effectiveness of transfer learning methods in brain tumor classification, demonstrating their potential impact on improving the diagnostic accuracy of medical image analysis. At the same time, they indicate the great potential of Vim in tumor classification tasks. The brain tumor classification framework proposed in this study is also applicable to all other medical image classification problems, with strong transferability.

## II. RELATED WORK

In this section, we discuss the current state of brain tumor classification using both traditional machine learning methods and deep learning approaches.

Traditional machine learning methods have been widely applied to brain tumor classification for many years. Support Vector Machine (SVM) is a classic supervised learning algorithm commonly used for classification and regression tasks [14]. In brain tumor classification, SVM constructs an optimal hyperplane in a high-dimensional space to separate tumors of different categories. The advantage of SVM lies in its strong generalization ability and suitability for small datasets. Lourzikene et al. employed SVM for feature extraction and classification of MRI image sequences, achieving high accuracy [15]. However, SVM is highly sensitive to parameter settings, such as kernel type and penalty parameters, necessitating complex parameter tuning.

Additionally, XGBoost is a gradient-boosted decision tree algorithm recognized for its computational efficiency and excellent classification performance [16]. In brain tumor classification, XGBoost iteratively constructs decision trees, using errors from previous iterations to improve the model's accuracy. Liu et al. applied XGBoost to brain tumor classification, demonstrating its advantage in handling high-dimensional data [17]. However, XGBoost also requires complex parameter tuning to achieve optimal results.

In recent years, the rise of deep learning has led to the development of CNNs capable of extracting deep features from MR images, resulting in successful classification outcomes. The Residual Network (ResNet) architecture proposed by He et al. has been widely adopted in image classification tasks, largely due to its use of skip connections, which help retain original input information [18]. In brain tumor classification, ResNet's deep feature extraction capabilities allow it to capture fine-grained details of different tumor types, thereby improving accuracy. Oladimeji and Ibitoye achieved notable results using ResNet for brain tumor classification [19]. Similarly, Sharif et al. utilized DenseNet201 for brain tumor classification. DenseNet's unique structure, characterized by its dense connectivity, allows the output of each layer to be directly connected to all subsequent layers [20]. This dense connectivity enhances feature reuse and mitigates the vanishing gradient problem, enabling effective classification even with limited training data.

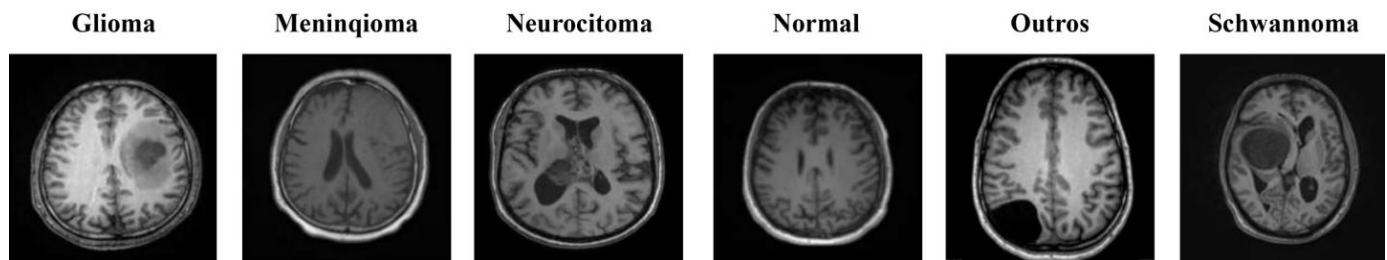

Fig. 1. Representative MRI images of different brain tumor classes and normal brain tissue. The figure shows examples of Glioma, Meningioma, Neurocytoma, Normal Brain, Other Brain Lesions, and Schwannoma, demonstrating the variability in appearance among different types of brain tumors and normal brain tissue.

Raza et al. employed the GoogLeNet architecture, which utilizes multiple Inception modules to capture multi-scale features from images via different convolutional kernels [21]. This approach reduces the number of parameters while maintaining the depth and complexity of the network, making it well-suited for processing medical images with intricate features. Raza validated the effectiveness of GoogLeNet in multi-class classification tasks, distinguishing tumors such as gliomas, meningiomas, and pituitary adenomas. Although CNN-based models have demonstrated significant success, they are inherently limited by their local receptive fields, which can hinder their ability to capture global information essential for accurate classification [22].

To address this limitation, researchers have begun introducing Transformer models into brain tumor classification. Compared to CNNs, the multi-head self-attention mechanism of Transformers allows the model to focus on different areas of the tumor, thereby capturing meaningful features from MRI images [23]. Ishak Pacal proposed an advanced deep learning approach using the Swin Transformer, which replaces the traditional MLP with a residual MLP to enhance feature extraction [24]. The Swin Transformer processes high-resolution images more effectively through a sliding window mechanism, reducing computational complexity while improving efficiency and accuracy. Another notable study by Khaniki et al. involved using the Vision Transformer (ViT) for brain tumor classification. Unlike CNNs, ViT segments the image into fixed-size patches and inputs these patches into a Transformer for processing [25]. Khaniki et al. incorporated optimization mechanisms such as feature calibration and selective cross-attention to improve feature integration, thereby enhancing classification accuracy [26]. However, despite their ability to capture long-range dependencies, Transformer-based models are characterized by high computational complexity, which limits their clinical applicability, particularly for small datasets prone to overfitting [27].

Given the high cost of obtaining medical image data and the laborious nature of annotation, transfer learning has emerged as an important research direction in medical image analysis. Transfer learning leverages pre-trained models on large-scale datasets (typically natural images) as a starting point, which are then fine-tuned on relatively small medical datasets [28]. This approach reduces the need for extensive annotated data and accelerates training. Hossain et al. employed six CNN-based models pre-trained on ImageNet and fine-tuned them on medical images to develop a new tumor classification model [29]. Similarly, Rehman et al. evaluated three CNN-based models—AlexNet, GoogLeNet, and VGGNet—using transfer learning for the classification of malignant brain tumors [30].

In summary, traditional CNN models have limitations in capturing global image information, which can prevent them from fully identifying subtle differences between tumors and surrounding tissues. Although Transformer models can handle long-range dependencies, their high computational complexity severely limits their clinical application. Recently, SSMs have garnered attention due to their efficiency and flexibility in sequence modeling [31]. The Mamba model is a novel type of SSM that effectively captures long-range dependencies through a selective mechanism while maintaining linear time complexity [32]. Mamba has demonstrated its superiority in classification tasks, especially in terms of computational resources and processing speed. It achieves fast inference under limited hardware conditions through efficient parameter optimization and a lightweight design, significantly reducing the time required for training and inference [11]. Vim, a variant of the Mamba model designed for visual tasks, is optimized for processing high-resolution image data [12]. Vim combines the accuracy of visual feature extraction with computational efficiency, making it highly effective in processing complex medical images. Its parallel processing and adaptive feature extraction mechanisms enhance its ability to capture edges and subtle structures, which is crucial in clinical environments.

This study explores the performance of the Vim model in multi-class brain tumor classification using single T1-weighted imaging input. This approach aims to reduce the dependency on multi-modal data and simplify data preprocessing, thereby enhancing the model's practicality and diagnostic efficiency. Currently, no studies have thoroughly examined the performance of Vim in MRI brain tumor multi-class classification. Our research fills this gap and proposes new methodologies for future medical image analysis.

To provide a comprehensive evaluation, we designed and conducted comparative experiments using traditional CNN models, Transformer-based models, and applied transfer learning. Both 2D and 3D network models were considered, and a 2D model was chosen for this experiment since 2D models can also handle 3D problems effectively and offer more options for transfer learning. Through this research, we aim to further enhance the accuracy and efficiency of brain tumor classification, thereby offering more reliable support for clinical diagnosis based on existing technologies.

## III. MATERIALS AND METHODS

### A. Dataset

This study employs a publicly accessible dataset of brain tumor images, commonly used for evaluating algorithms related to multi-class tumor classification [1]. The dataset comprises a total of 4,449 authentic MRI slices acquired in the axial plane, including T1-weighted imaging (T1WI), contrast-enhanced T1-weighted imaging (CE-T1WI), and T2-weighted imaging (T2WI). These images were categorized into six distinct classes of brain tumor pathologies, as illustrated in Fig. 1.

1) **Glioma**
   Tumors originating from glial cells, which play a supportive role in maintaining normal neural function within the brain and spinal cord. Gliomas are classified into several subtypes based on the cell type of origin and malignancy level: astrocytoma, ganglioglioma, glioblastoma, oligodendroglioma, and ependymoma. High-grade gliomas, such as glioblastoma multiforme, are more invasive with a poor prognosis, while low-grade gliomas grow more slowly and generally respond better to treatment.

2) **Meningioma**
   Tumors arising from the meninges, which are the protective membranes surrounding the brain and spinal cord. Meningiomas are usually benign but can be further classified into subtypes such as grade I (benign), atypical, anaplastic, and transitional based on cellular characteristics and malignancy potential.

3) **Neurocitoma**
   Tumors arising from neuroglial cells that provide support to neurons in the brain and spinal cord. Neurocytomas are primarily categorized as either central neurocytoma or extraventricular neurocytoma, depending on their anatomical location.

4) **Normal Brain Tissue**
   This category includes normal brain tissue, without presence of tumor of any kind.

5) **Outros**
   This category includes conditions other than tumor such as abscesses, cysts, and miscellaneous encephalopathies. Although not classified as typical brain tumors, these lesions can present together with brain tumors and thus require appropriate diagnostic and therapeutic intervention.

6) **Schwannoma**
   Tumors originating from the cells of the neural sheath, which form the protective covering around nerve fibers. Schwannomas include types such as acoustic, vestibular, and trigeminal schwannomas.

Given the diversity of brain tumor types, accurate classification is crucial for the development of targeted intervention strategies and treatment protocols, underscoring the clinical significance of precise tumor typing.

Despite the dataset includes multiple imaging modalities, CE-T1WI is often regarded as optimal for diagnosing brain tumors. However, CE-T1WI is not always available in clinical practice. Therefore, this study focuses on the most used imaging modality in clinical settings, T1WI, as input. This subset comprises 1,554 images. For transfer learning, pre-trained models based on the ImageNet dataset require an input image channel count of three. Since MRI images are grayscale, we replicated the grayscale values across three channels. The specific division of the dataset for training, validation, and testing is shown in Table I. The training set contains 80% of the images, while the validation and testing sets each contain 10%.

TABLE I
SPLIT OF THE DATASET INTO TRAINING, VALIDATION, AND TESTING SETS. THE TRAINING SET CONTAINS 80% OF THE IMAGES, WHILE THE VALIDATION AND TESTING SETS EACH CONTAIN 10%. THE TABLE PRESENTS THE NUMBER OF IMAGES FOR EACH TUMOR TYPE, DIVIDED ACROSS THE THREE SETS

|  | Train | Val | Test | Total |
|---|---|---|---|---|
| Glioma | 371 | 46 | 46 | 463 |
| Meninqioma | 276 | 34 | 35 | 345 |
| Normal | 217 | 27 | 28 | 272 |
| Neurocitoma | 135 | 17 | 17 | 169 |
| Outros | 121 | 15 | 16 | 152 |
| Schwannoma | 122 | 15 | 16 | 153 |
| Total | 1242 | 154 | 158 | 1554 |

### B. Model

Transfer learning is a technique where knowledge gained from a model trained on one task is transferred to a different but related task, thereby improving the learning performance in the target domain. In this study, we first evaluated the performance of five classic pre-trained models on the brain tumor classification task and subsequently assessed the novel Vim model. All six models were initialized with pre-trained weights from the ImageNet dataset, and transfer learning was applied for tumor classification:

1) **ViT**
   The self-attention mechanism in ViT is one of its key innovations, providing a novel way to process image data compared to traditional CNNs by capturing global dependencies among different parts of the input. Specifically, ViT calculates relationships between each image patch and all other patches to obtain contextual information, allowing for an enhanced understanding of the image content through global information [33]. In the self-attention mechanism, each patch undergoes a linear transformation to produce query, key, and value vectors. The model calculates the attention scores between these vectors, indicating the significance of each patch in understanding the global image context.

---

[1] https://www.kaggle.com/datasets/fernando2rad/brain-tumor-mri-images-17-classes

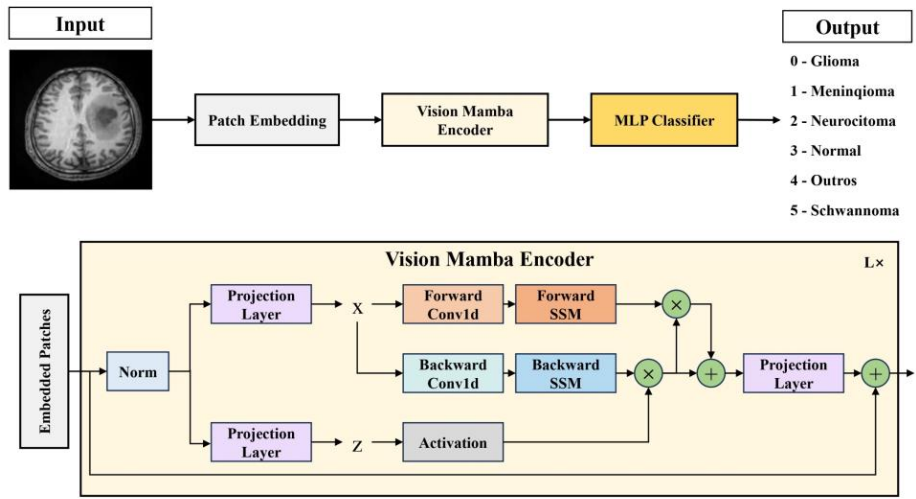

Fig. 2. Overview of the Vision Mamba model. The model begins by splitting the input MRI image into patches, followed by projecting these patches into embedded tokens. The sequence of tokens is then processed by the Vision Mamba encoder. Finally, multi-layer perceptron (MLP) layers are used for classification.

By taking a weighted average of value vectors, ViT generates a context-aware representation that allows each patch to incorporate information from others. Unlike CNNs, which primarily capture local features. ViT can process the entire image at once, which is especially important when dealing with large datasets to identify complex patterns and details.

2) **Swin Transformer**

Swin Transformer is an advanced visual Transformer model that optimizes high-resolution image processing through hierarchical self-attention and a shifting window mechanism [34]. It divides images into non-overlapping local windows, applying self-attention within each window to efficiently capture local features. The window-shifting strategy then facilitates communication between different regions, thereby improving the model's global understanding of the image.

3) **EfficientNet-B0**

EfficientNet-B0 introduces an efficient network scaling method compared to traditional approaches. The model uses the compound scaling principle to uniformly scale all dimensions, including depth, width, and resolution, using predefined scaling coefficients. EfficientNet-B0 employs MobileNetV2-inspired inverted bottleneck convolution (MBConv) blocks and depth-wise separable convolutions, minimizing parameters and computational load while maintaining accuracy [35]. The model also uses squeeze-and-excitation optimization to enhance the network's representational capacity by recalibrating channel-level feature responses.

4) **Inception-V3**

InceptionV3, part of the Inception model series developed by Google, is known for its efficiency and accuracy in image classification tasks. The core idea of InceptionV3 is to optimize feature extraction by applying convolution operations with different kernel sizes (e.g., 1x1, 3x3, 5x5) in parallel [36]. This approach allows for multi-scale feature extraction without significantly increasing computational complexity. 1x1 convolutions are used for dimensionality reduction before applying larger kernels to further reduce computational costs.

5) **ResNet-50**

ResNet-50 is a deep CNN architecture from the ResNet family, proposed by Microsoft Research [37]. Its main innovation is residual learning through skip connections, which alleviates the vanishing gradient problem in deep networks. ResNet-50 has 50 layers, including residual blocks that allow the model to learn residual functions, thereby facilitating deeper feature extraction. These skip connections ensure that even very deep networks can maintain performance without degradation.

6) **Vision Mamba**

The Vim model is a novel deep learning architecture specifically designed for efficient visual representation learning. This model is based on SSM, particularly adapted from the Mamba model initially developed for natural language processing tasks involving long sequence modeling [12]. Vim extends this approach to the visual domain, effectively addressing unique challenges such as positional sensitivity and the need for global context in image data.

Vim overcomes these challenges by introducing a bidirectional state space model, which allows it to process both forward and backward information in the sequence, thereby capturing a comprehensive global context of data dependencies. As a pure SSM-based visual backbone network, Vim treats images as sequences of flattened patches and compresses visual representations through a bidirectional selective state space model without relying on the self-attention mechanism, which is commonly used in ViT. This approach overcomes the quadratic computational complexity associated with Transformers. By introducing positional embeddings, Vim can better understand complex spatial relationships in visual data.

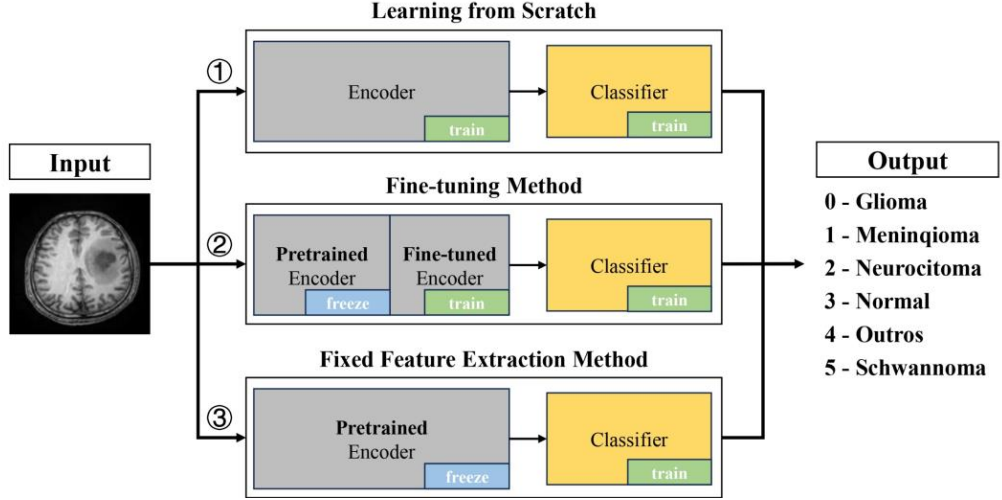

Fig. 3. Strategies of model training. Freeze means that the weights remain unchanged during training. Train implies that the weights are updated during training. The encoder consists of three architectures, i.e., CNN, Transformer, and Mamba.

As illustrated in Fig. 2, the Vim model first transforms 2D images into a sequence of patches, which are then mapped into vectors with added positional encoding, like the ViT model. A class token is also included to represent the entire patch sequence. These vectors are fed into the Vim encoder, and the final output of the class token is passed to a multi-layer perceptron (MLP) to generate the final prediction. The original Mamba module was designed for 1D data, which is unsuitable for visual tasks requiring spatial perception. Vim adapts the Mamba approach by combining bidirectional sequence modeling to suit visual data. Specifically, the Vim module normalizes the sequence, linearly projects it into two components (x and z), and then processes x in both forward and backward directions. For each direction, a 1D convolution is first applied, followed by processing in the SSM module. The resulting forward and backward outputs ($y_{forward}$ and $y_{backward}$) are gated by z and combined to obtain the final output from the Vim module.

The described models, including both well-established architectures and the novel Vim model, provide a diverse set of capabilities to address the challenges of brain tumor classification. The inclusion of classic CNN-based architectures and Transformer-based models, along with the unique SSM-based Vim, allows us to comprehensively assess their efficacy in tumor classification. This combination of models aims to highlight the effectiveness of different approaches and to explore novel methods like Vim for enhancing diagnostic accuracy in medical image analysis.

## IV. EXPERIMENTS AND RESULTS

### A. Experimental Design

To tailor the original models for the specific domain of multi-class brain tumor classification, we modified their penultimate layers. The original fully connected (FC) layer in each transfer learning architecture was replaced with a new FC layer with an output dimension of six, corresponding to the six brain tumor classes under investigation. The Adam optimizer was used for optimization, with cross-entropy loss as the loss function—an appropriate choice for multi-class classification tasks. The learning rate was dynamically adjusted during training, incorporating scheduled reductions at predetermined phases to mitigate overfitting risks. Additionally, early stopping was used as a regularization technique to halt training when performance on the validation set plateaued for a specific number of epochs.

First, we trained the network model from scratch; then, we loaded pre-trained weights and fine-tuned the layers closer to the output. Finally, we froze the encoder layers of the pre-trained model and trained only the fully connected layers for classification. Our findings indicated that the model achieved superior training efficacy while maintaining computational efficiency when the transfer learning involved unfreezing the last 20 layers only. For the Vim model, however, all layers were unfrozen after loading pre-trained weights to maximize its adaptability. The settings were consistent for the other models. The overall workflow is depicted in Fig. 3.

We employed a comprehensive set of metrics to evaluate model performance, including accuracy, precision, recall, f1-score, specificity, sensitivity, and the area under the receiver operating characteristic curve (AUC). Classification results were visualized using confusion matrices, which facilitated the calculation of accuracy, precision, recall (or sensitivity), specificity, and f1-score. Additionally, AUC was derived from class probability estimates provided by each model.

We also conducted an analysis to explore the relationship between model complexity, represented by the number of parameters and the number of floating-point operations (FLOPs), and the accuracy metric. This analysis aimed to provide insights into the trade-offs between model complexity and performance. Our evaluation was designed to be thorough, comparing the merits and drawbacks of each model. This comprehensive assessment enabled us to identify the most suitable model for the brain tumor multi-classification task, considering predictive accuracy, computational efficiency, and generalizability.

TABLE II
PERFORMANCE METRICS OF ALL MODELS ON THE TUMOR DATASET. * DENOTES THE TRANSFER LEARNING RESULTS OF THE CORRESPONDING MODEL

| model | accuracy | precision | recall | f1-score | specificity | sensitivity | AUC |
|---|---|---|---|---|---|---|---|
| swintransformer[34] | 0.92 | 0.93 | 0.92 | 0.92 | 0.98 | 0.92 | 0.99 |
| swintransformer* | 0.92 | 0.93 | 0.92 | 0.92 | 0.98 | 0.92 | 0.99 |
| vit[33] | 0.90 | 0.90 | 0.90 | 0.90 | 0.97 | 0.90 | 0.99 |
| vit* | 0.96 | 0.96 | 0.96 | 0.96 | 0.99 | 0.96 | 0.99 |
| efficientnet-b0[35] | 0.95 | 0.95 | 0.95 | 0.95 | 0.99 | 0.95 | 0.99 |
| efficientnet-b0* | 0.97 | 0.98 | 0.97 | 0.97 | 0.99 | 0.97 | 1.00 |
| inception-v3[36] | 0.96 | 0.96 | 0.96 | 0.96 | 0.99 | 0.96 | 1.00 |
| inception-v3* | 0.97 | 0.97 | 0.97 | 0.97 | 0.99 | 0.97 | 1.00 |
| resnet50[37] | 0.97 | 0.97 | 0.97 | 0.97 | 0.99 | 0.97 | 1.00 |
| resnet50* | 0.98 | 0.98 | 0.98 | 0.98 | 0.99 | 0.98 | 1.00 |
| visionmamba[12] | 0.92 | 0.93 | 0.92 | 0.92 | 0.98 | 0.92 | 0.99 |
| visionmamba* | 1.00 | 1.00 | 1.00 | 1.00 | 1.00 | 1.00 | 1.00 |

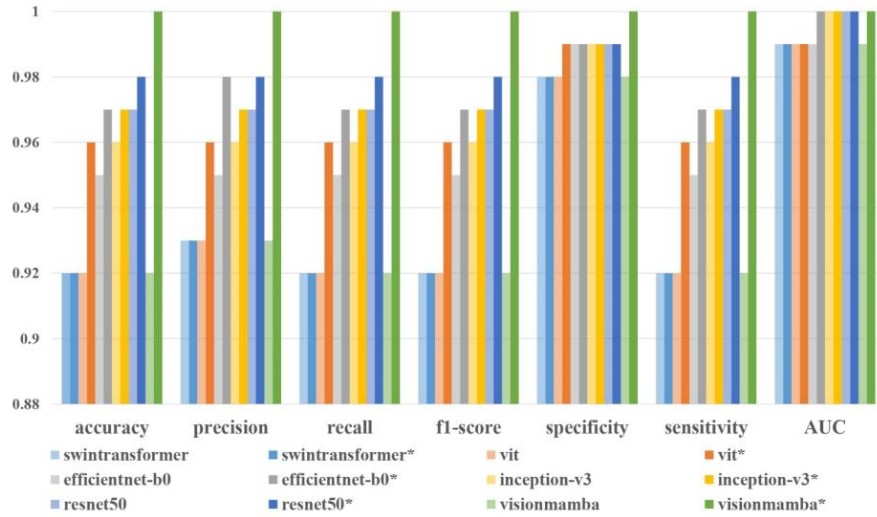

Fig. 4. Accuracy, precision, recall, f1-score, specificity, sensitivity and AUC of all models using the tumor dataset without and with transfer learning. The same color represents the same model, with lighter shades indicating the results from training from scratch, and darker shades indicating the results of transfer learning.

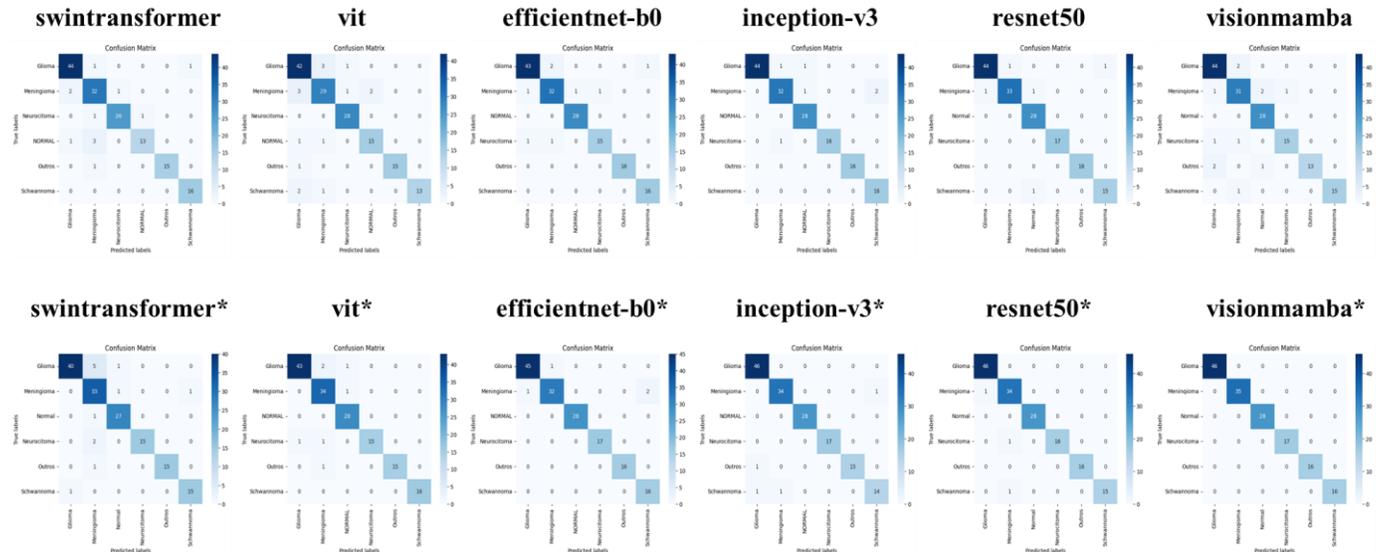

Fig. 5. Confusion matrices for all models. The top row displays the results of models trained from scratch, while the bottom row shows the results of models trained using transfer learning. Each confusion matrix presents the classification performance for the six tumor categories.

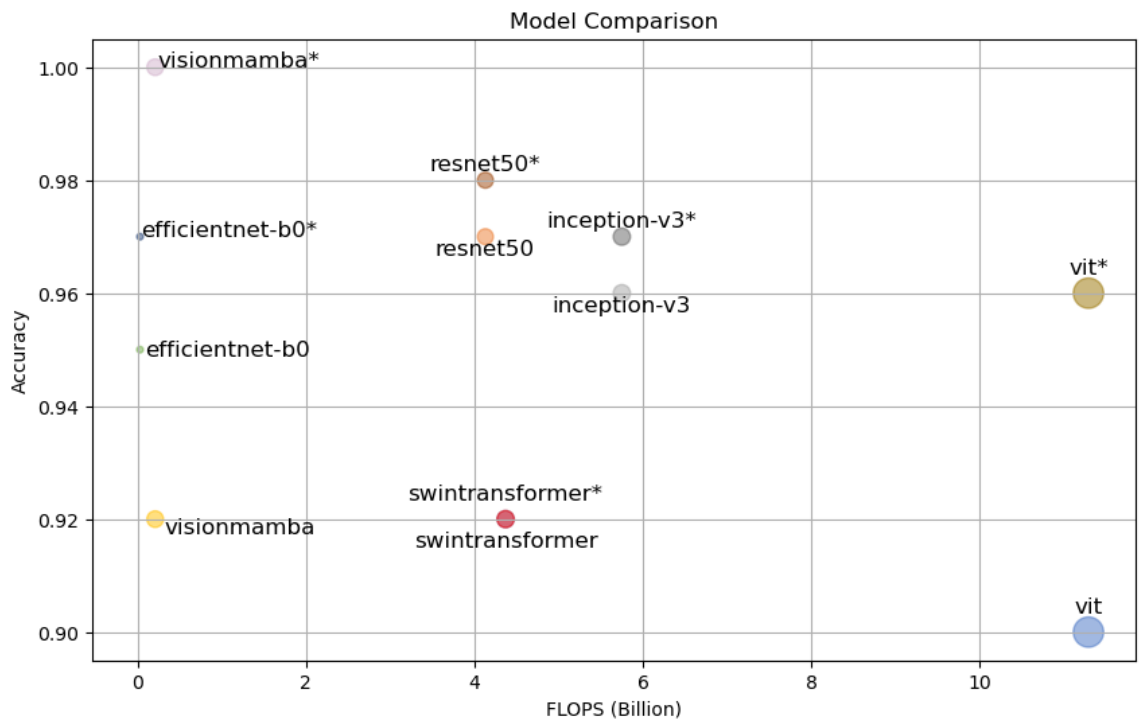

Fig. 6. Comparison of model performance based on computational complexity (FLOPs in billions) versus accuracy. Each point represents a model, with the x-axis indicating the number of floating-point operations (FLOPs) and the y-axis showing overall accuracy. Models with an asterisk (*) indicate the use of transfer learning. The size of each point is proportional to the number of model parameters, highlighting the trade-off between model complexity and accuracy.

## B. Results and Analysis

In this part, we compare the performance of five classic transfer learning models—both trained from scratch and with pre-training—alongside the results of the Vim model, which was also trained from scratch and with pre-training. The specific evaluation results for each metric are presented in Table II and Fig. 4. The test accuracy of the Swin Transformer, ViT, EfficientNet-B0, Inception-V3, and ResNet50 models all exceeded 0.90, while the Vim model achieved a test accuracy of 0.92. In the transfer learning setting, all models achieved classification accuracies above 0.95, except for the Swin Transformer, which achieved 0.90. For almost all models, transfer learning led to a noticeable improvement in performance. The Vim model achieved the best classification result, correctly identifying all test samples with an accuracy of 100%. The confusion matrices illustrating the classification evaluation results for each model are shown in Fig. 5.

The evaluation results clearly indicate that the transfer learning approach significantly enhances classification performance. Additionally, during transfer learning, we observed that using pre-trained weights allowed the models to achieve optimal classification performance more quickly, requiring fewer training epochs and resulting in higher training efficiency. We also analyzed the relationship between the number of model parameters, computational complexity in terms of FLOPs, and model classification accuracy, as illustrated in Fig. 6. In the figure, the horizontal axis represents the computational complexity in FLOPs, and the vertical axis represents the model's classification accuracy. The size of each circle indicates the number of model parameters, with larger circles representing models with more parameters. From the figure, we can observe that both the EfficientNet-B0 and Vim models exhibit low computational complexity while maintaining high efficiency and accuracy. The pre-trained Vim model achieved a classification accuracy of 100%, highlighting it as a lightweight and efficient model.

Moreover, the ResNet50 and Inception-V3 models, regardless of whether transfer learning was applied, both achieved accuracies above 0.95. However, the computational complexity of these models is significantly higher compared to the Vim model.

Based on the experimental results, the use of transfer learning consistently resulted in superior performance compared to models trained from scratch, demonstrating higher efficiency and better generalization capabilities. The Vim model, as a novel Mamba-based network, performed exceptionally well in the tumor classification task, with its transfer learning results reaching an impressive 100% accuracy on the independent test set. This remarkable performance reflects the substantial potential of the Vim architecture in tumor classification applications. Additionally, Vim is a lightweight and efficient model, ensuring high accuracy while being more practical for future clinical applications due to its lower computational requirements compared to other models. The combination of high accuracy and computational efficiency makes the Vim model a promising candidate for real-world medical use.

## V. DISCUSSION

Medical imaging encompasses a wide range of heterogeneity, making accurate image detection essential for the interpretation of medical data. In clinical practice, MRI and CT imaging are commonly used for brain tumor identification, with MRI frequently preferred for tumor detection and classification. This study utilized deep learning techniques based on MRI imaging to classify different types of brain tumors, demonstrating that various standard CNNs can achieve high accuracy in this task. Specifically, we employed mainstream transfer learning models, including Swin Transformer, Vision Transformer, EfficientNet, InceptionV3, and ResNet50, to perform multi-class classification of six types of brain tumors: Glioma, Meningioma, Neurocitoma, Normal Brain Tissue, Other Brain Lesions, Schwannoma.

In addition, we compared the performance of these mainstream transfer learning models to the Vision Mamba model in multi-class classification of brain tumors. To explore the benefits of transfer learning based on pre-trained weights, we trained each model three times: once from scratch, once using pre-trained weights with fine-tuning of the later network layers, and once using pre-trained weights where only the fully connected layers for output were trained. The models were evaluated based on metrics such as accuracy, precision, recall, f1-score, specificity, sensitivity, and area under the receiver operating characteristic curve (AUC). Our experiments showed that transfer learning significantly improved the accuracy and efficiency of classification tasks, with the Vision Mamba model achieving a classification accuracy of 100% on the independent test set. Additionally, we analyzed the relationship between the number of model parameters, FLOPs, and classification accuracy, finding that Vision Mamba is both lightweight and efficient.

These results demonstrate the effectiveness of transfer learning for brain tumor classification and underscore the significant potential of the Vision Mamba model in multi-class tumor classification tasks. Transfer learning with pre-training provides better generalization compared to models specifically trained for a single task, and it converges more quickly and efficiently during training. Conversely, models trained from scratch are more challenging to optimize and require strict control over hyperparameters. The framework proposed in this study, based on transfer learning and the Vision Mamba model for brain tumor classification, is broadly applicable to other medical imaging classification problems.

However, this study also has limitations. First, the dataset used in this study is relatively limited in size. Furthermore, according to the World Health Organization (WHO), there are numerous types of brain tumors, whereas our current dataset only includes a small number of categories. As a result, we cannot guarantee that the model's performance will remain consistent when expanded to include additional tumor types. An important area for future research is the creation of a dataset that includes images from different MRI scanners and covers more brain tumor categories, thereby increasing the diversity and volume of data for classification tasks. Additionally, incorporating other clinical imaging modalities, such as CT, could further improve classification precision.

## VI. CONCLUSION AND FUTURE WORK

This study introduces the Mamba-based Vim model to brain tumor classification, using transfer learning to optimize performance within medical imaging. By reducing data dependency and computational complexity, Vim offers a novel framework that could benefit a range of clinical imaging tasks beyond brain tumor classification. Its state-space model architecture bypasses the need for traditional attention mechanisms while maintaining high accuracy, making it a promising, lightweight solution for real-world clinical applications. Furthermore, the framework proposed in this study for brain tumor classification, based on transfer learning and the Vision Mamba model, is broadly applicable to other medical imaging classification problems.

In future work, data augmentation techniques could be employed to expand the dataset size, thereby improving model robustness. We also intend to further study training strategies for transfer learning to better apply these pre-trained models to downstream tasks, such as image classification. Exploring ways to optimize the Vision Mamba model's architecture to make it even more lightweight and efficient will also be a key focus, enhancing its utility in clinical applications. Future research could also focus on different stages of various diseases, including early prevention, clinical treatment, and prognosis prediction, utilizing deep learning technology to improve clinical management and outcomes.

## REFERENCES AND FOOTNOTES

[1] M. J. van den Bent *et al.*, "Primary brain tumours in adults," *The Lancet*, vol. 402, no. 10412, pp. 1564–1579, Oct. 2023, doi: 10.1016/S0140-6736(23)01054-1.
[2] "Focusing on brain tumours and brain metastasis," *Nat Rev Cancer*, vol. 20, no. 1, pp. 1–1, Jan. 2020, doi: 10.1038/s41568-019-0232-7.
[3] S. Bauer, R. Wiest, L.-P. Nolte, and M. Reyes, "A survey of MRI-based medical image analysis for brain tumor studies," *Phys. Med. Biol.*, vol. 58, no. 13, p. R97, Jun. 2013, doi: 10.1088/0031-9155/58/13/R97.
[4] A. Lerner, M. A. Mogensen, P. E. Kim, M. S. Shiroishi, D. H. Hwang, and M. Law, "Clinical Applications of Diffusion Tensor Imaging," *World Neurosurgery*, vol. 82, no. 1, pp. 96–109, Jul. 2014, doi: 10.1016/j.wneu.2013.07.083.
[5] M. Sonka, S. K. Tadikonda, and S. M. Collins, "Knowledge-based interpretation of MR brain images," *IEEE Transactions on Medical Imaging*, vol. 15, no. 4, pp. 443–452, Aug. 1996, doi: 10.1109/42.511748.
[6] J. Amin, M. Sharif, M. Yasmin, and S. L. Fernandes, "A distinctive approach in brain tumor detection and classification using MRI," *Pattern Recognition Letters*, vol. 139, pp. 118–127, Nov. 2020, doi: 10.1016/j.patrec.2017.10.036.
[7] N. M. Dipu, S. A. Shohan, and K. M. A. Salam, "Deep Learning Based Brain Tumor Detection and Classification," in *2021 International Conference on Intelligent Technologies (CONIT)*, Jun. 2021, pp. 1–6. doi: 10.1109/CONIT51480.2021.9498384.
[8] J. Lin *et al.*, "CKD-TransBTS: Clinical Knowledge-Driven Hybrid Transformer With Modality-Correlated Cross-Attention for Brain Tumor Segmentation," *IEEE Transactions on Medical Imaging*, vol. 42, no. 8, pp. 2451–2461, Aug. 2023, doi: 10.1109/TMI.2023.3250474.
[9] A. Hatamizadeh *et al.*, "UNETR: Transformers for 3D Medical Image Segmentation," in *2022 IEEE/CVF Winter Conference on Applications of Computer Vision (WACV)*, Jan. 2022, pp. 1748–1758. doi: 10.1109/WACV51458.2022.00181.
[10] A. Gu, K. Goel, and C. Ré, "Efficiently Modeling Long Sequences with Structured State Spaces," arXiv.org. Accessed: Sep. 29, 2024. [Online]. Available: https://arxiv.org/abs/2111.00396v3
[11] A. Gu and T. Dao, "Mamba: Linear-Time Sequence Modeling with Selective State Spaces," arXiv.org. Accessed: Aug. 11, 2024. [Online]. Available: https://arxiv.org/abs/2312.00752v2
[12] L. Zhu, B. Liao, Q. Zhang, X. Wang, W. Liu, and X. Wang, "Vision Mamba: Efficient Visual Representation Learning with Bidirectional State Space Model," arXiv.org. Accessed: Sep. 23, 2024. [Online]. Available: https://arxiv.org/abs/2401.09417v2
[13] H. H. Sultan, N. M. Salem, and W. Al-Atabany, "Multi-Classification of Brain Tumor Images Using Deep Neural Network," *IEEE Access*, vol. 7, pp. 69215–69225, 2019, doi: 10.1109/ACCESS.2019.2919122.
[14] M. Awad and R. Khanna, "Support Vector Machines for Classification," in *Efficient Learning Machines: Theories, Concepts, and Applications for*

*Engineers and System Designers*, M. Awad and R. Khanna, Eds., Berkeley, CA: Apress, 2015, pp. 39–66. doi: 10.1007/978-1-4302-5990-9_3.

[15] "Performance Evaluation of Feature Extraction and SVM for Brain Tumor Detection Using MRI Images | IIETA." Accessed: Sep. 23, 2024. [Online]. Available: https://iieta.org/journals/ts/paper/10.18280/ts.410426

[16] T. Chen and C. Guestrin, "XGBoost: A Scalable Tree Boosting System," in *Proceedings of the 22nd ACM SIGKDD International Conference on Knowledge Discovery and Data Mining*, in KDD '16. New York, NY, USA: Association for Computing Machinery, Aug. 2016, pp. 785–794. doi: 10.1145/2939672.2939785.

[17] M. Jha, R. Gupta, and R. Saxena, "A framework for in-vivo human brain tumor detection using image augmentation and hybrid features," *Health Inf Sci Syst*, vol. 10, no. 1, p. 23, Aug. 2022, doi: 10.1007/s13755-022-00193-9.

[18] K. He, X. Zhang, S. Ren, and J. Sun, "Deep Residual Learning for Image Recognition," arXiv.org. Accessed: Sep. 23, 2024. [Online]. Available: https://arxiv.org/abs/1512.03385v1

[19] O. O. Oladimeji and A. O. J. Ibitoye, "Brain tumor classification using ResNet50-convolutional block attention module," *Applied Computing and Informatics*, vol. ahead-of-print, no. ahead-of-print, Jan. 2023, doi: 10.1108/ACI-09-2023-0022.

[20] M. I. Sharif, M. A. Khan, M. Alhussein, K. Aurangzeb, and M. Raza, "A decision support system for multimodal brain tumor classification using deep learning," *Complex Intell. Syst.*, vol. 8, no. 4, pp. 3007–3020, Aug. 2022, doi: 10.1007/s40747-021-00321-0.

[21] A. Raza *et al.*, "A Hybrid Deep Learning-Based Approach for Brain Tumor Classification," *Electronics*, vol. 11, no. 7, Art. no. 7, Jan. 2022, doi: 10.3390/electronics11071146.

[22] Y. Xie *et al.*, "Convolutional Neural Network Techniques for Brain Tumor Classification (from 2015 to 2022): Review, Challenges, and Future Perspectives," *Diagnostics*, vol. 12, no. 8, Art. no. 8, Aug. 2022, doi: 10.3390/diagnostics12081850.

[23] Z. Liu, Q. Lv, Z. Yang, Y. Li, C. H. Lee, and L. Shen, "Recent progress in transformer-based medical image analysis," *Computers in Biology and Medicine*, vol. 164, p. 107268, Sep. 2023, doi: 10.1016/j.compbiomed.2023.107268.

[24] I. Pacal, "A novel Swin transformer approach utilizing residual multi-layer perceptron for diagnosing brain tumors in MRI images," *Int. J. Mach. Learn. & Cyber.*, Mar. 2024, doi: 10.1007/s13042-024-02110-w.

[25] G. Boesch, "Vision Transformers (ViT) in Image Recognition - 2024 Guide," viso.ai. Accessed: Sep. 23, 2024. [Online]. Available: https://viso.ai/deep-learning/vision-transformer-vit/

[26] M. A. L. Khaniki, A. Golkarieh, and M. Manthouri, "Brain Tumor Classification using Vision Transformer with Selective Cross-Attention Mechanism and Feature Calibration," arXiv.org. Accessed: Sep. 23, 2024. [Online]. Available: https://arxiv.org/abs/2406.17670v1

[27] Y. Dai and Y. Gao, "TransMed: Transformers Advance Multi-modal Medical Image Classification," arXiv.org. Accessed: Sep. 23, 2024. [Online]. Available: https://arxiv.org/abs/2103.05940v1

[28] S. Asif, W. Yi, Q. U. Ain, J. Hou, T. Yi, and J. Si, "Improving Effectiveness of Different Deep Transfer Learning-Based Models for Detecting Brain Tumors From MR Images," *IEEE Access*, vol. 10, pp. 34716–34730, 2022, doi: 10.1109/ACCESS.2022.3153306.

[29] S. Hossain, A. Chakrabarty, T. R. Gadekallu, M. Alazab, and Md. J. Piran, "Vision Transformers, Ensemble Model, and Transfer Learning Leveraging Explainable AI for Brain Tumor Detection and Classification," *IEEE Journal of Biomedical and Health Informatics*, vol. 28, no. 3, pp. 1261–1272, Mar. 2024, doi: 10.1109/JBHI.2023.3266614.

[30] A. Rehman, S. Naz, M. I. Razzak, F. Akram, and M. Imran, "A Deep Learning-Based Framework for Automatic Brain Tumors Classification Using Transfer Learning," *Circuits Syst Signal Process*, vol. 39, no. 2, pp. 757–775, Feb. 2020, doi: 10.1007/s00034-019-01246-3.

[31] J. Jiang, F. Deng, G. Singh, M. Lee, and S. Ahn, "Slot State Space Models," arXiv.org. Accessed: Sep. 23, 2024. [Online]. Available: https://arxiv.org/abs/2406.12272v5

[32] Y. Suo, Z. Ding, and T. Zhang, "The Mamba Model: A Novel Approach for Predicting Ship Trajectories," *Journal of Marine Science and Engineering*, vol. 12, no. 8, Art. no. 8, Aug. 2024, doi: 10.3390/jmse12081321.

[33] D. Alexey, "An image is worth 16x16 words: Transformers for image recognition at scale," *arXiv* preprint arXiv: 2010.11929, 2020.

[34] Z. Liu, Y. Lin, C. Cao et al., "Swin transformer: Hierarchical vision transformer using shifted windows," *Proceedings of the IEEE/CVF international conference on computer vision*. 2021: 10012-10022.

[35] M. Tan, "Efficientnet: Rethinking model scaling for convolutional neural networks," *arXiv* preprint arXiv: 1905.11946, 2019.

[36] S. M. Sam, K. Kamardin, N. N. A. Sjarif, et al., "Offline signature verification using deep learning convolutional neural network (CNN) architectures GoogLeNet inception-v1 and inception-v3," *Procedia Computer Science*, 2019, 161: 475-483.

[37] O. Russakovsky, J. Deng, H. Su et al., "Imagenet large scale visual recognition challenge," *International journal of computer vision*, 2015, 115: 211-252.